\documentclass[11pt]{article}
\usepackage{graphicx} 
\usepackage{amsmath,amsfonts,amssymb,amsthm}
\usepackage{geometry}
\usepackage{times}
\usepackage{titlesec}
\usepackage{enumitem}
\usepackage{fancyhdr}
\usepackage{hyperref}
\usepackage{multirow} 
\usepackage{algorithm}
\usepackage{algpseudocode}
\titleformat{\section}{\large\bfseries}{\thesection.}{1em}{}
\titleformat{\subsection}{\normalsize\bfseries}{\thesubsection.}{1em}{}
\titleformat{\subsubsection}{\normalsize\bfseries\itshape}{\thesubsubsection.}{1em}{}

\theoremstyle{definition}

\theoremstyle{remark}

\title{\textbf{State Estimation Using Particle Filtering in Adaptive Machine Learning Methods: Integrating Q-Learning and NEAT Algorithms with Noisy Radar Measurements}}
\author{
Wonjin Song\\
\textit{Department of Mathematics, Florida State University, Tallahassee, FL}\\
\texttt{ws20cc@fsu.edu}
\and
Feng Bao\\
\textit{Department of Mathematics, Florida State University, Tallahassee, FL}\\
\texttt{bao@math.fsu.edu}
}

\date{}

\begin{document}

\maketitle

\hrule  


\begin{abstract}
Reliable state estimation is essential for autonomous systems operating in complex, noisy environments. Classical filtering approaches, such as the Kalman filter, can struggle when facing nonlinear dynamics or non-Gaussian noise, and even more flexible particle filters often encounter sample degeneracy or high computational costs in large-scale domains. Meanwhile, adaptive machine learning techniques, including Q-learning and neuroevolutionary algorithms such as NEAT, rely heavily on accurate state feedback to guide learning; when sensor data are imperfect, these methods suffer from degraded convergence and suboptimal performance. In this paper, we propose an integrated framework that unifies particle filtering with Q-learning and NEAT to explicitly address the challenge of noisy measurements. By refining radar-based observations into reliable state estimates, our particle filter drives more stable policy updates (in Q-learning) or controller evolution (in NEAT), allowing both reinforcement learning and neuroevolution to converge faster, achieve higher returns or fitness, and exhibit greater resilience to sensor uncertainty. Experiments on grid-based navigation and a simulated car environment highlight consistent gains in training stability, final performance, and success rates over baselines lacking advanced filtering. Altogether, these findings underscore that accurate state estimation is not merely a preprocessing step, but a vital component capable of substantially enhancing adaptive machine learning in real-world applications plagued by sensor noise.
\end{abstract}

\section{Introduction}

State estimation is a fundamental problem in control and data assimilation, where the goal is to determine the true state of a dynamical system from noisy or incomplete observations. Classic approaches like the Kalman filter \cite{kalman1960,kalman1961new} offer closed-form solutions under linear and Gaussian assumptions, providing computational efficiency in moderate-dimensional problems. However, real-world systems often break these assumptions, causing extended and unscented Kalman filters \cite{julier2000unscented} to deteriorate in performance when facing strong nonlinearity or non-Gaussian noise.

To address more general scenarios, particle filters \cite{gordon1993novel,doucet2000sequential,kitagawa1998self} were introduced as simulation-based methods to approximate the Bayesian posterior distribution. By propagating and reweighting an ensemble of particles, these filters can capture nonlinear effects and complex distributions. Despite their conceptual appeal, particle filters can struggle in high-dimensional settings due to issues such as sample degeneracy and high computational cost \cite{snyder2008obstacles,vanleeuwen2010nonlinear}, motivating extensive research into alternative or hybrid strategies.

One family of methods tackles filtering through partial or stochastic differential equations (PDE/SPDE) formulations, such as the Zakai or Kushner-Stratonovich equations \cite{zakai1969optimal,bao2014hybrid,bao2015forward,bao2016first,bao2017adaptive}, which provide a mathematically rigorous framework for posterior evolution. Though SPDE-based methods can achieve stable performance, they often entail heavy computational demands, making them less suitable for real-time or large-scale applications \cite{bao2020data,bao2021adaptive}. Ensemble Kalman filters \cite{evensen1994sequential,houtemaker1998data}, variational data assimilation \cite{lorenc1986analysis}, and various ensemble-variational hybrids have also been developed to handle higher dimensions while retaining some of the advantages of Monte Carlo sampling and Bayesian inference. In recent years, score-based diffusion filters have emerged as a promising approach, using generative diffusion models \cite{ho2020denoising,song2019generative} to approximate complex, high-dimensional probability densities. Efforts by Bao et al.\ \cite{bao_united,bao_score} incorporate these diffusion-based ideas into novel filtering frameworks that can handle nonlinear dynamical systems and unknown parameters with greater accuracy.

Another important challenge arises when model parameters must be estimated jointly with the state. 
Traditional approaches, such as an augmented state formulation in ensemble Kalman filtering, treat parameters as extra components in the state vector, potentially amplifying dimensionality issues \cite{evensen2009data, anderson2001ensemble}. Meanwhile, Kitagawa's self-organizing filter \cite{kitagawa1998self} also addresses joint parameter and state estimation by augmenting the state-space model, although without strictly relying on 
ensemble Kalman methods. Recent work integrates diffusion-based score filters with a direct parameter filtering strategy \cite{bao2020unitedfilter}, thereby mitigating computational bottlenecks in joint state-parameter estimation for high-dimensional or nonlinear dynamical systems. Further advances include refined ensemble score methods for large-scale tracking \cite{bao2024ensemble} and robust particle filtering via drift homotopy  \cite{li2022particle}, illustrating the growing interest in alternatives to naive augmentation.

In parallel with these filtering advances, adaptive machine learning methods have gained increasing attention in control and decision-making. Reinforcement learning (RL) and neuroevolutionary methods both fit under this umbrella, as they learn from interactions with an environment or from iterative evaluations of candidate policies. On the RL side, value-based algorithms expand Q-learning \cite{watkins1992q} into deep neural networks, as in Deep Q-Networks (DQN) \cite{mnih2015human}, which handle complex discrete-action tasks. Rainbow DQN \cite{hessel2018rainbow} adds multiple enhancements for stability and sample efficiency. Policy gradient methods, exemplified by Proximal Policy Optimization (PPO) \cite{schulman2017proximal} and Soft Actor Critic (SAC) \cite{haarnoja2018soft}, have become popular in continuous control tasks, while Twin Delayed DDPG (TD3) \cite{fujimoto2018addressing} and Trust Region Policy Optimization (TRPO) \cite{schulman2015trust} address overestimation bias and constraints on policy updates. These algorithms continue to evolve through model-based RL approaches (e.g., MuZero \cite{schrittwieser2020mastering}, Dreamer \cite{hafner2020dreamer}) and offline RL or Transformer-based methods \cite{kumar2020conservative,kostrikov2021offline,chen2021decision}, broadening their application to multi-task or fixed-dataset scenarios.

Nevertheless, most RL approaches assume they have sufficiently accurate state information to guide learning updates, an assumption that often fails in sensor-driven environments subject to noise, occlusions, or abrupt changes. Bridging robust state estimation with RL thus remains critical in real-world applications. Neuroevolutionary methods, meanwhile, explore a different corner of adaptive learning by employing evolutionary search to optimize neural network weights and topologies. The NeuroEvolution of Augmenting Topologies (NEAT) algorithm \cite{stanley2002evolving} adaptively expands network structures over generations. Extensions \cite{stanley2004competitive,risi2019deep,miikkulainen2019evolving} address scalability and increasingly complex tasks, yet they too rely on accurate feedback signals to guide the evolutionary process.

In this paper, we focus on particle filtering, a proven approach for nonlinear and non-Gaussian estimation, and show how it can be combined with two adaptive learning methods—Q-learning and NEAT—to improve decision-making under uncertainty. By refining noisy radar-based sensor measurements, the particle filter produces more reliable state estimates, enabling faster convergence and better performance in tasks such as grid-based pathfinding and autonomous navigation. These results emphasize that accurate state estimation is not merely a preprocessing step, but a core component that substantially enhances adaptive control and learning strategies in challenging domains.

In the remainder of this paper, Section~\ref{sec:problem} introduces the grid-based navigation and car navigation tasks, highlighting the challenges of sensor noise and motivating the need for robust state estimation. Section~\ref{sec:method} then presents our integrated approach, detailing the particle filtering technique and its integration with Q-learning and NEAT. Section~\ref{sec:experimental} outlines the simulation environments and experimental parameters, while Section~\ref{sec:results} reports extensive numerical results that demonstrate the benefits of accurate state estimation in both tasks. Finally, Section~\ref{sec:conclusion} concludes the paper by summarizing key findings and discussing possible directions for future work.

\section{Problem Setting}
\label{sec:problem}

In many real-world control tasks, an agent must operate under uncertainty, receiving noisy measurements of its true state. We investigate two navigation problems under noise: a grid-based environment solved by Q-learning (Section~\ref{subsec:grid}), and a simulated car environment solved by NEAT (Section~\ref{subsec:car}). Although Q-learning fits naturally into a Markov Decision Process (MDP) formulation and NEAT does not strictly require an MDP, both methods depend on accurate state estimation when noise is present.

\subsection{Grid-based Navigation (Q-learning)}
\label{subsec:grid}

Consider an agent with state $s \!=\! (x,y)\in[0,12]\times[0,12]$. At time $t$, it selects an action $(v,\theta)$ from a finite set $\mathcal{A}$, where $v$ is a (bounded) speed and $\theta$ is one of several discrete angles. The state evolves as
\[
s_{t+1} \;=\; s_t \;+\; v 
\begin{pmatrix}
\cos(\theta_t) \\[6pt] \sin(\theta_t)
\end{pmatrix}
\;+\;\eta_t,
\quad 
\eta_t \;\sim\;\mathcal{N}\!\bigl(0,\sigma^2 I\bigr).
\]
A reward function $\mathcal{R}(s,a,s')$ grants positive rewards for following a target path and penalizes boundary collisions or stalling. Formally, this defines a Markov Decision Process, making Q-learning a natural choice. However, if the agent only observes noisy measurements $z_t$, reliable filtering of $s_t$ becomes essential for accurate Q-learning updates.

\subsection{Car Navigation (NEAT)}
\label{subsec:car}

We also study a car environment with state $s$ possibly including position $(x,y)$, orientation $\theta$, and velocity $v$. Control inputs (steering angle, acceleration) are real-valued. Radar or other sensors yield noisy observations,
\[
z_t = s_t + \eta_t, 
\quad \eta_t \sim \mathcal{N}(0,\sigma^2 I),
\]
which complicates the agent’s knowledge of its own state. NeuroEvolution of Augmenting Topologies (NEAT) evolves neural network controllers $\pi_\theta(s)$ to maximize a fitness function linked to distance traveled, checkpoints reached, or obstacle avoidance. Although NEAT does not use an MDP framework, it remains sensitive to inaccurate state estimates induced by sensor noise.

\subsection{Challenges in Noisy State Estimation}
\label{subsec:challenges}

In both the grid-based (Q-learning) and car-navigation (NEAT) scenarios, the true state $s_t$ is never directly observed; instead, each agent receives
\[
z_t \;=\; s_t + \eta_t.
\]
Kalman-type filters assume linear or mildly nonlinear dynamics under Gaussian conditions but may fail in more general settings. Particle filtering, by approximating the posterior $p(s_t \mid z_{1:t})$ with weighted samples, can handle greater nonlinearity at the expense of higher computational cost. In the following section, we describe our integrated approach that employs particle filtering to refine these noisy measurements, enabling both Q-learning and NEAT to learn or evolve effective solutions despite the noise.

\section{Methodology}
\label{sec:method}

We now present our integrated framework, which combines particle filtering with Q-learning and NEAT to address navigation under noise. Section~\ref{subsec:filter} outlines the particle filter used to estimate $s_t$ from $z_t$. Sections~\ref{subsec:Qlearning} and \ref{subsec:NEAT} explain how Q-learning and NEAT each incorporate the filtered state.

\subsection{Particle Filtering for State Estimation}
\label{subsec:filter}

Let $s_t$ be the (hidden) state at time $t$ and $z_t$ the noisy observation:
\[
z_t = s_t + \eta_t, 
\quad \eta_t \sim \mathcal{N}(0,\sigma^2 I).
\]
We approximate the posterior $p(s_t \mid z_{1:t})$ using $N$ particles $\{ s_t^{(i)}, w_t^{(i)}\}_{i=1}^N$. The key steps are:

\begin{enumerate}
    \item \textbf{Initialization:} Sample $s_0^{(i)} \sim p(s_0)$ and set $w_0^{(i)}=1/N$.
    \item \textbf{Prediction:} Assume 
    \[
    s_{t+1} = f(s_t,a_t) + \omega_t, \quad \omega_t \sim \mathcal{N}(0,\sigma^2 I).
    \]
    For each particle $s_t^{(i)}$, predict
    \[
    s_{t+1}^{(i)} = f\bigl(s_t^{(i)},a_t\bigr) + \omega_t^{(i)}.
    \]
    \item \textbf{Weight Update:} Observe $z_{t+1}$, set
    \[
    w_{t+1}^{(i)} \;\propto\; w_t^{(i)}\,\exp\!\Bigl(-\|z_{t+1}-s_{t+1}^{(i)}\|^2 / (2\sigma^2)\Bigr).
    \]
    Normalize so $\sum_i w_{t+1}^{(i)}=1$.
    \item \textbf{Resampling:} Resample $\{s_{t+1}^{(i)}, w_{t+1}^{(i)}\}$ if weights become too uneven.
\end{enumerate}

A refined state estimate is $\hat{s}_{t+1} = \sum_{i=1}^N w_{t+1}^{(i)} s_{t+1}^{(i)}$, which replaces raw measurements in Q-learning or NEAT.

\subsection{Integration with Q-Learning}
\label{subsec:Qlearning}

Q-learning learns an optimal action-value function $Q^*(s,a)$ satisfying
\[
Q^*(s,a) 
\;=\; \mathbb{E}\Bigl[r_t + \gamma \max_{a'} Q^*(s_{t+1},a') \,\Big|\; s,a\Bigr].
\]
We update $Q(\hat{s}_t,a_t)$ by
\[
Q(\hat{s}_t,a_t) \;\leftarrow\; Q(\hat{s}_t,a_t)
+\alpha \Bigl[r_t + \gamma \max_{a'}Q(\hat{s}_{t+1},a') - Q(\hat{s}_t,a_t)\Bigr].
\]
Here, $\hat{s}_t$ is the particle filter’s estimate of the agent’s true position or configuration, reducing noise-driven instabilities in the Q-update.

\subsection{Integration with NEAT}
\label{subsec:NEAT}

NeuroEvolution of Augmenting Topologies (NEAT) evolves neural network controllers $\pi_\theta(s)$ through an evolutionary algorithm. Each controller’s performance is measured by
\[
F(\theta) = \sum_{t=0}^T \gamma^t\,\mathcal{R}\bigl(\hat{s}_t,\pi_\theta(\hat{s}_t)\bigr),
\]
where $\hat{s}_t$ is again the filtered state estimate at time $t$. By feeding $\hat{s}_t$ (rather than a noisy measurement) into each controller, we obtain more reliable fitness evaluations, guiding NEAT to discover controllers that genuinely handle the environment rather than artifacts of sensor noise.

\subsection{Algorithmic Flow}

Algorithm~\ref{alg:integrated} summarizes the integrated approach. At each timestep, the particle filter refines $z_t$ into $\hat{s}_t$, which is then used by either Q-learning or NEAT, depending on the experiment.

\begin{algorithm}[h]
\caption{Integrated Particle Filtering with Q-learning and NEAT}
\label{alg:integrated}
\begin{algorithmic}[1]
\State \textbf{Input:} Particle count $N$, Q-table $Q(s,a)$, NEAT population, initial prior $p(s_0)$.
\For{t=0,1,2,\ldots}
    \State Observe noisy measurement $z_t$.
    \State \textbf{Particle Filter Update:} 
      \For{i=1 to N}
        \State $s_t^{(i)} \leftarrow f(s_{t-1}^{(i)}, a_{t-1}) + \omega_{t-1}^{(i)};$
        \State $w_t^{(i)} \propto w_{t-1}^{(i)} \exp\bigl(-\|z_t - s_t^{(i)}\|^2 / (2\sigma^2)\bigr).$
      \EndFor
      \State Normalize and possibly resample $\{s_t^{(i)}, w_t^{(i)}\}.$
      \State $\hat{s}_t \leftarrow \sum_{i=1}^N w_t^{(i)} s_t^{(i)}.$
    \State \textbf{If Q-learning:}
      \State Observe reward $r_t,$ choose action $a_t$ (e.g.\ $\epsilon$-greedy).
      \State Update $Q(\hat{s}_t,a_t)$ via the Bellman relation.
    \State \textbf{If NEAT:}
      \State For each controller $\pi_\theta$, compute fitness increment using $\hat{s}_t.$
      \State Evolve NEAT population (selection, crossover, mutation).
\EndFor
\State \textbf{Output:} Final Q-table or NEAT controllers.
\end{algorithmic}
\end{algorithm}

\section{Simulation and Experimental Setup}
\label{sec:experimental}

We now describe the concrete numerical parameters used in each environment. Complete hyperparameter lists are provided in Appendices~A and~B.

\subsection{Q-Learning Experiment}
\label{subsec:grid_exp}

The grid-based domain $[0,12]\times[0,12]$ is discretized into 51 points per axis. The agent always starts at $(x_0,y_0) = (2.8,2.8)$ in the lower corner of the grid and attempts to follow a wave-shaped path for $30{,}000$ episodes. The action space has $8$ directions and a bounded speed from $0.4$ to $1.4.$ Radar noise has variance $0.07^2.$ Particle filtering uses $N=500$ samples. Q-learning uses learning rate $\alpha=0.001,$ discount $\gamma=0.999,$ and an $\epsilon$ schedule decaying from $1.0$ to $10^{-5}.$ The exact reward shaping and boundary penalties appear in Appendix~A.

\subsection{NEAT Experiment}
\label{subsec:car_exp}

We adapt our car simulation environment from an open-source tutorial by CheesyAI \cite{CheesyAIGithub}, incorporating modifications for our particle filter integration. The state $s$ includes $(x,y,\theta,v)$, and the agent applies real-valued steering and acceleration. Noisy radar measurements use $\sigma^2$ varied across experiments. NEAT evolves a population of $30$ individuals for $20$ generations, where the fitness includes distance traveled and checkpoints reached. We also test different levels of sensor noise to observe how the particle filter influences the evolutionary process. Appendix~B lists population-level mutation rates and other NEAT parameters.

\subsection{Implementation Notes}

The experiments were implemented in Python. All random-number generation relies on \texttt{numpy.random} and Python’s \texttt{random} module with their default seeding mechanisms, meaning that each run may yield slightly different outcomes. (If strict reproducibility is required, one can set a fixed seed near the start of the script via \texttt{np.random.seed(\dots)} and \texttt{random.seed(\dots)}.)

Both the grid-based and car-navigation tasks share the same particle filtering procedure; only the system dynamics function \(\,f(s,a)\) and the nature of reward or fitness evaluation differ. The entire codebase, including hyperparameters, is provided in the Appendices to facilitate replication of our experiments (with the understanding that results will vary slightly unless a fixed seed is explicitly set).

\section{Results and Analysis}
\label{sec:results}

This section presents empirical outcomes from our two main tasks: grid-based navigation (Q-learning) and car navigation (NEAT). We evaluate the effect of integrating particle filtering on learning stability, convergence, and final performance. Hyperparameters, noise levels, and episode/generation counts follow the descriptions in Section~\ref{sec:experimental} and Appendices~A--B.

\subsection{Q-Learning Experiment}

We measure the Q-learning agent’s performance primarily via average reward per episode and final success rate. Let $R_i$ denote the total return in episode $i$, with $N$ total episodes. The average reward is
\[
\bar{R} = \frac{1}{N} \sum_{i=1}^{N} R_i.
\]
We also compute a sliding-window average (over 50 episodes) to observe convergence behavior.

\begin{figure}[h]
    \centering
    \begin{minipage}{0.48\textwidth}
        \centering
        \includegraphics[width=\textwidth]{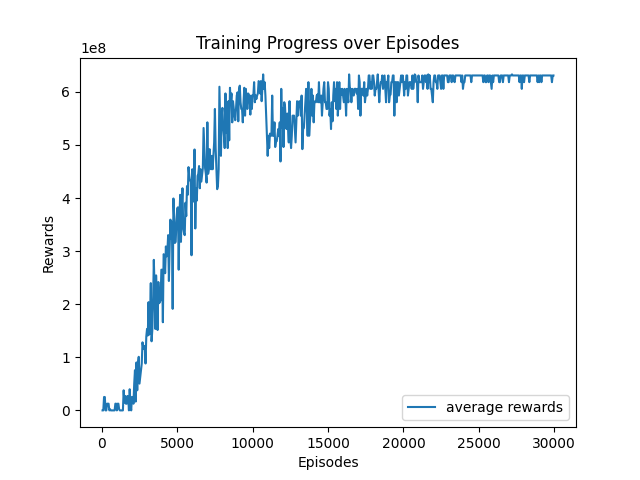}
        \caption{Average rewards with particle filtering.}
        \label{fig:particle_avg_reward}
    \end{minipage}
    \hfill
    \begin{minipage}{0.48\textwidth}
        \centering
        \includegraphics[width=\textwidth]{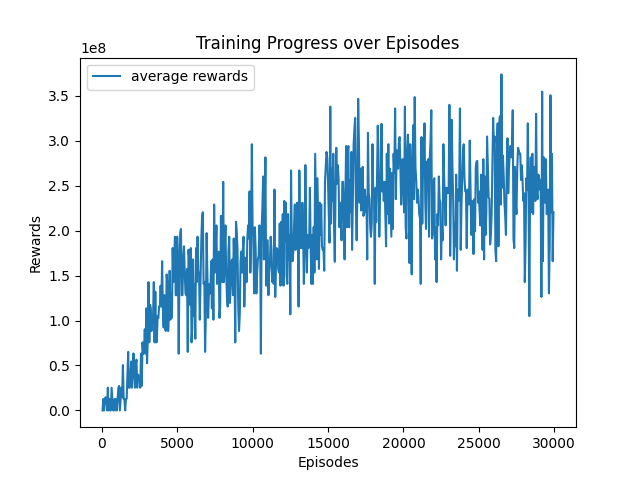}
        \caption{Average rewards without particle filtering.}
        \label{fig:noisy_avg_reward}
    \end{minipage}
\end{figure}

\paragraph{Average Reward and Convergence.}
Figures~\ref{fig:particle_avg_reward} and \ref{fig:noisy_avg_reward} compare the learning curves with (left) and without (right) particle filtering, each plotted over 30{,}000 episodes. The particle-filtering agent attains consistently higher average rewards and stabilizes more rapidly. In contrast, the baseline case (no filtering) exhibits more variability and lower asymptotic rewards, indicating that accurate state estimates are crucial for effective Q-learning updates.

\paragraph{Stability and Coefficient of Variation.}
To assess training stability, we track the coefficient of variation (CV), defined as the ratio of the standard deviation to the mean reward within specific phases of training. Table~\ref{tab:mean_variance_cv_comparison} shows that the non-filtered agent exhibits a higher CV for both early and late phases, reflecting greater volatility. The agent with particle filtering achieves a sharply reduced CV in later training, underscoring more reliable convergence.

\begin{table}[h!]
\caption{Comparison of Mean Reward, Variance, and CV in Early vs.\ Late Training (Q-learning).}
\label{tab:mean_variance_cv_comparison}
\centering
\begin{tabular}{|c|c|c|c|c|c|c|}
\hline
\multirow{2}{*}{\textbf{Phase}} & \multicolumn{3}{c|}{\textbf{With Filtering}} & \multicolumn{3}{c|}{\textbf{Without Filtering}} \\
\cline{2-7}
 & Mean & Variance & CV & Mean & Variance & CV \\
\hline
Early (0--10k) & $3.05\times10^{8}$ & $4.49\times10^{16}$ & 0.69 & $1.10\times10^{8}$ & $4.33\times10^{15}$ & 0.60 \\ 
Late (20k--30k)& $6.25\times10^{8}$ & $9.54\times10^{13}$ & 0.016 & $2.44\times10^{8}$ & $2.42\times10^{15}$ & 0.20 \\
\hline
\end{tabular}
\end{table}

\paragraph{Final Success Rate.}
We define a success as reaching the final target within the episode. Table~\ref{tab:success_rate} indicates that particle filtering boosts the success rate from about $33\%$ to about $67\%$, demonstrating the benefits of robust state estimation in navigating a wave-shaped path under noisy measurements.

\begin{table}[h]
    \centering
    \caption{Success rate after training (Q-learning).}
    \label{tab:success_rate}
    \begin{tabular}{|c|c|c|}
        \hline
        & \textbf{With Filtering} & \textbf{Without Filtering} \\
        \hline
        Success Rate & 66.98\% & 32.85\% \\
        \hline
    \end{tabular}
\end{table}

These results confirm that particle filtering stabilizes Q-learning in noisy environments, yielding higher rewards and a substantially improved success rate.

\subsection{Reward Stabilization Analysis}

\begin{figure}[h]
    \centering
    \begin{minipage}[b]{0.49\textwidth}
        \centering
        \includegraphics[width=\textwidth]{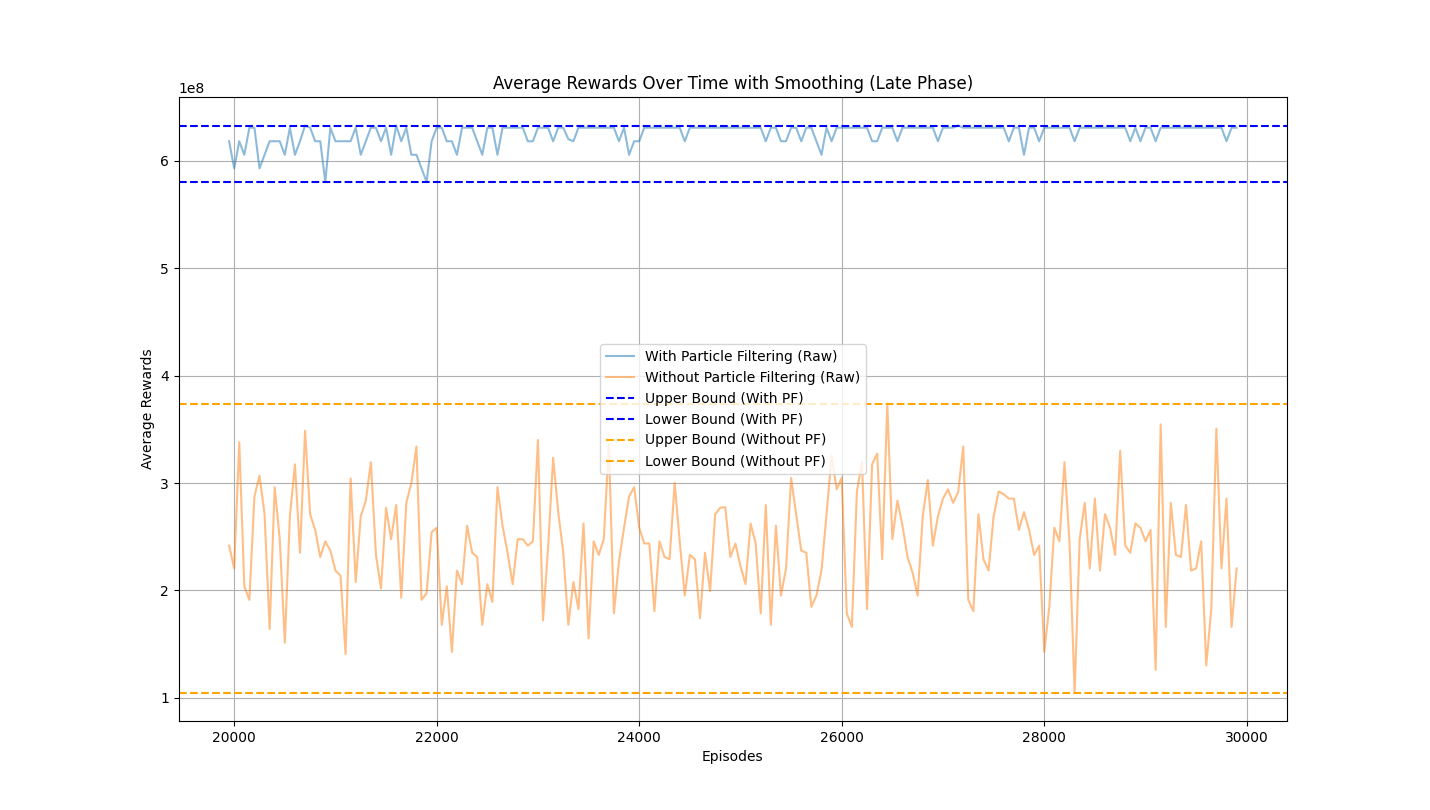}
        \caption{Average rewards (late phase) with upper/lower bounds, illustrating more stable performance under filtering.}
        \label{fig:avg_rewards_stabilization}
    \end{minipage}\hfill
    \begin{minipage}[b]{0.49\textwidth}
        \centering
        \includegraphics[width=\textwidth]{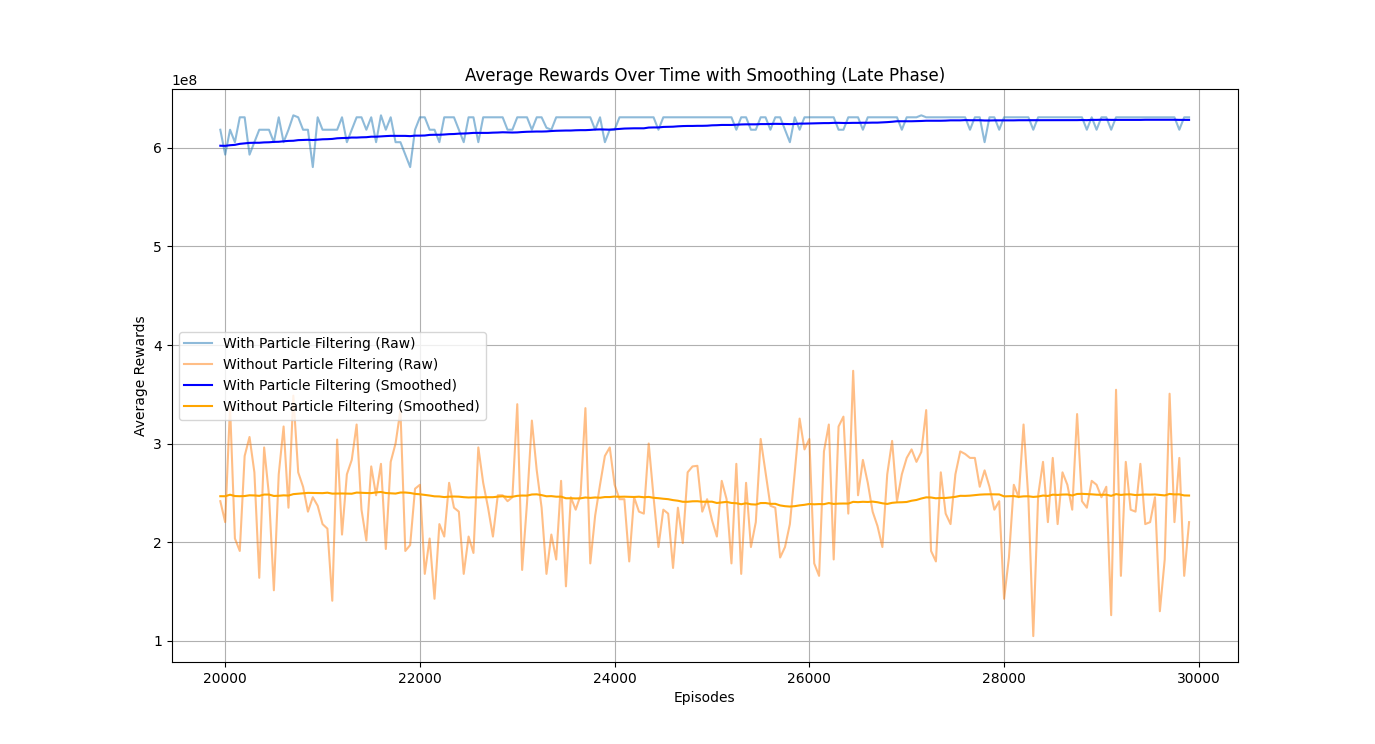}
        \caption{Smoothed average rewards (late phase). Particle filtering case converges more stably toward high returns.}
        \label{fig:avg_rewards_convergence}
    \end{minipage}
\end{figure}

Figures~\ref{fig:avg_rewards_stabilization} and \ref{fig:avg_rewards_convergence} further illustrate late-phase learning dynamics. In Figure~\ref{fig:avg_rewards_stabilization}, the agent with filtering exhibits a tighter reward range, while the baseline’s rewards fluctuate more widely. Smoothed curves in Figure~\ref{fig:avg_rewards_convergence} show faster convergence to near-optimal values when filtering is used.

\begin{figure}[h]
    \centering
    \begin{minipage}{0.49\textwidth}
        \centering
        \includegraphics[width=\textwidth]{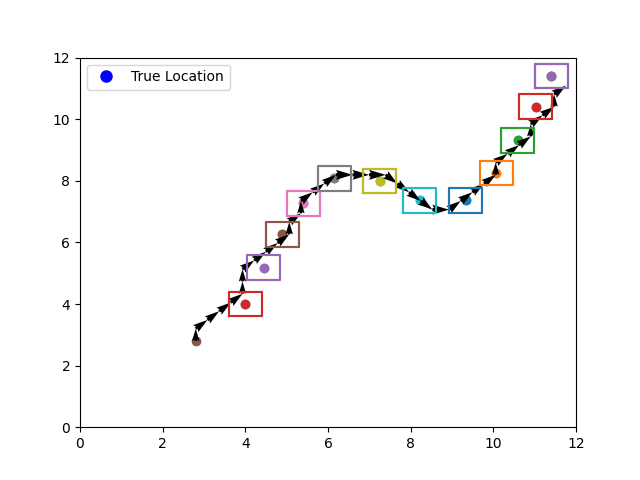}
        \caption{Near-optimal path (with filtering).}
        \label{fig:with_pf}
    \end{minipage}
    \hfill
    \begin{minipage}{0.49\textwidth}
        \centering
        \includegraphics[width=\textwidth]{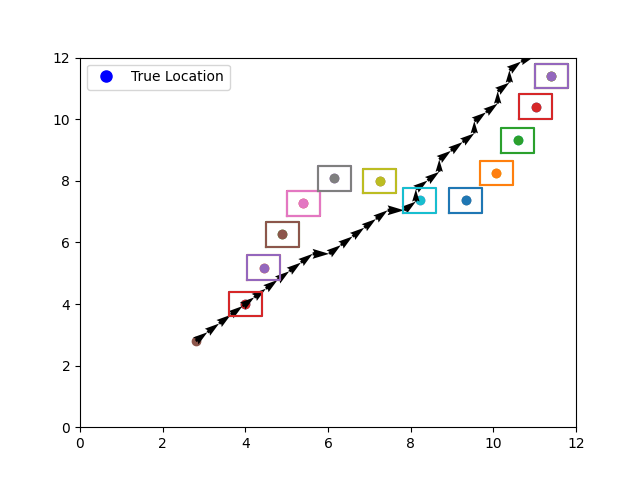}
        \caption{Poor path (no filtering).}
        \label{fig:without_pf}
    \end{minipage}
\end{figure}

\paragraph{Final Path Visualization.}
Figure~\ref{fig:with_pf} (left) shows a near-optimal path to the target when the agent uses particle filtering, while Figure~\ref{fig:without_pf} (right) illustrates a meandering path failing to reach the target in the baseline. This further underscores the impact of better state estimates on final policy quality.

\subsection{NEAT Experiment}

We now turn to the NEAT-based car-navigation task, measuring \emph{fitness} (distance traveled, checkpoints, etc.) across generations under different noise levels. We compare runs with and without particle filtering to highlight the effect on evolutionary outcomes.

\begin{figure}[h]
    \centering
    \begin{minipage}[b]{0.495\textwidth}
        \centering
        \includegraphics[width=\textwidth]{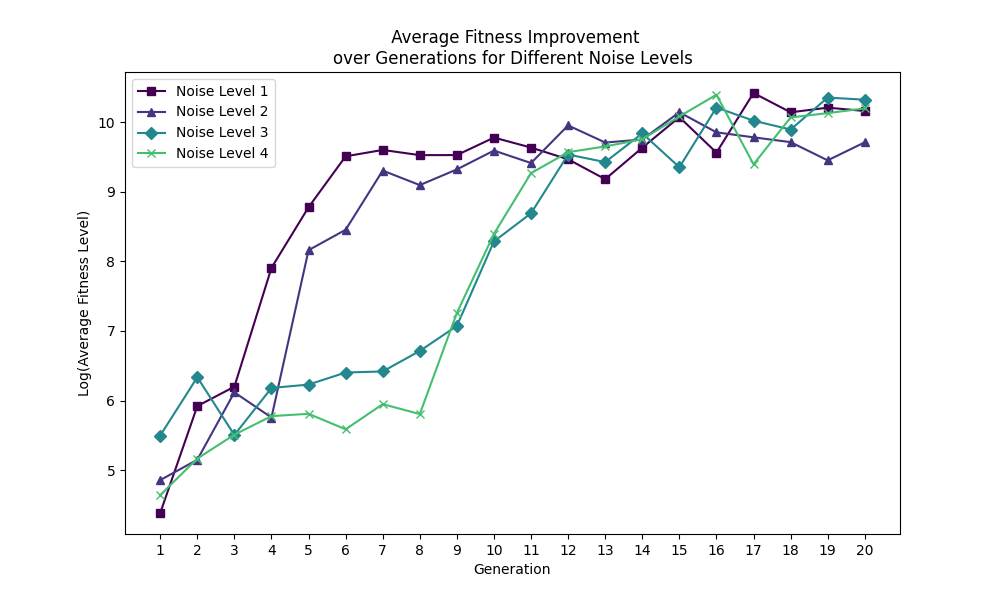}
        \caption{Average fitness with filtering.}
        \label{fig:average_fitness_with_particle_filter}
    \end{minipage}
    \hfill
    \begin{minipage}[b]{0.495\textwidth} 
        \centering
        \includegraphics[width=\textwidth]{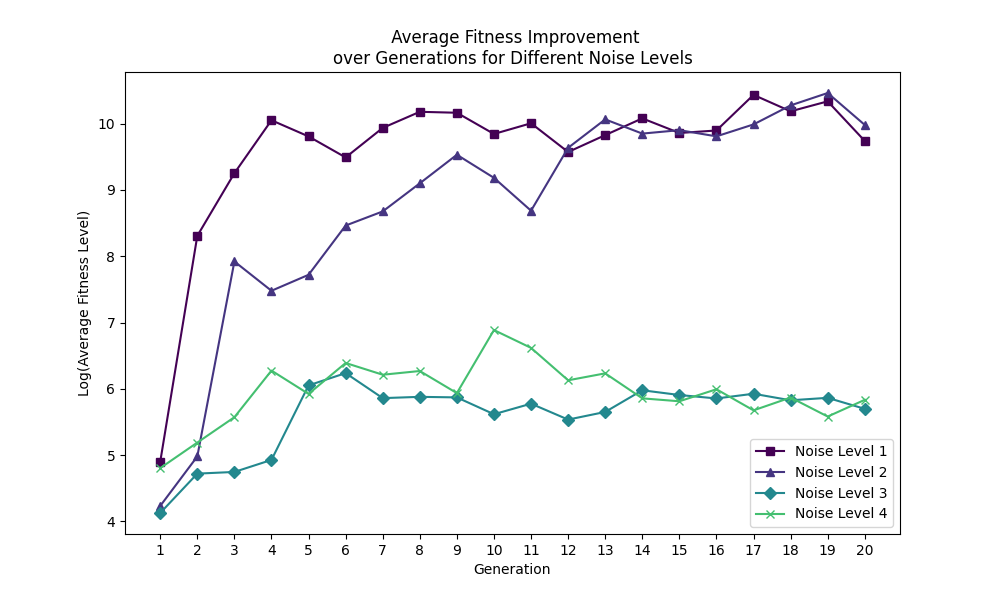}
        \caption{Average fitness without filtering.}
        \label{fig:average_fitness_no_particle_filter}
    \end{minipage}
\end{figure}

\paragraph{Average and Best Fitness.}
Figures~\ref{fig:average_fitness_with_particle_filter} and \ref{fig:average_fitness_no_particle_filter} track average fitness over 20 generations. Particle filtering leads to faster improvement and higher final fitness, reflecting more accurate evaluations of candidate controllers. Without filtering, fitness often plateaus or regresses at higher noise levels, suggesting sensitivity to flawed state inputs.

\begin{figure}[h]
    \centering
    \begin{minipage}[b]{0.495\textwidth}
        \centering
        \includegraphics[width=\textwidth]{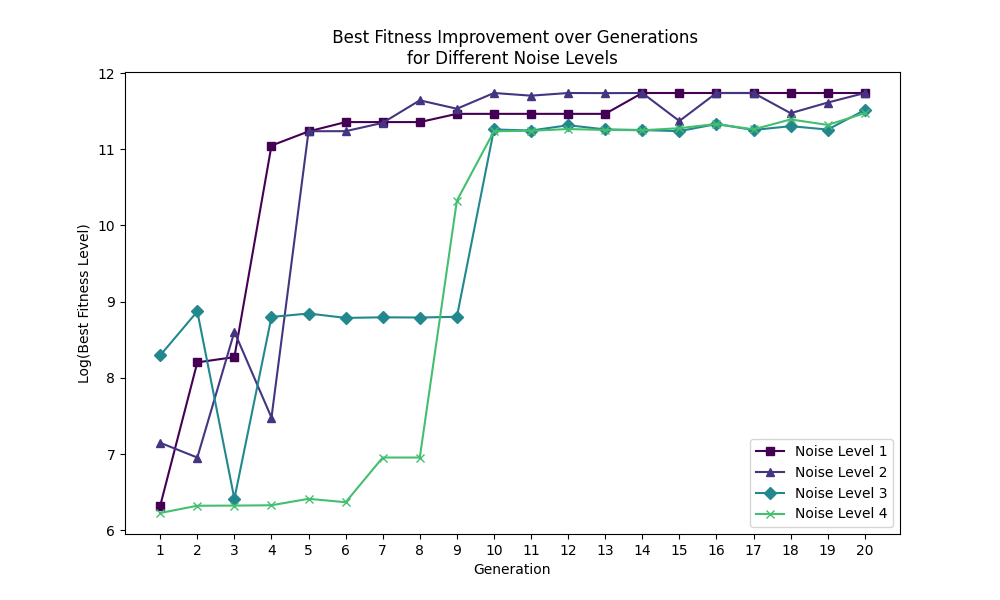}
        \caption{Best fitness (with filtering).}
        \label{fig:best_fitness_with_particle_filter}
    \end{minipage}
    \hfill
    \begin{minipage}[b]{0.495\textwidth}
        \centering
        \includegraphics[width=\textwidth]{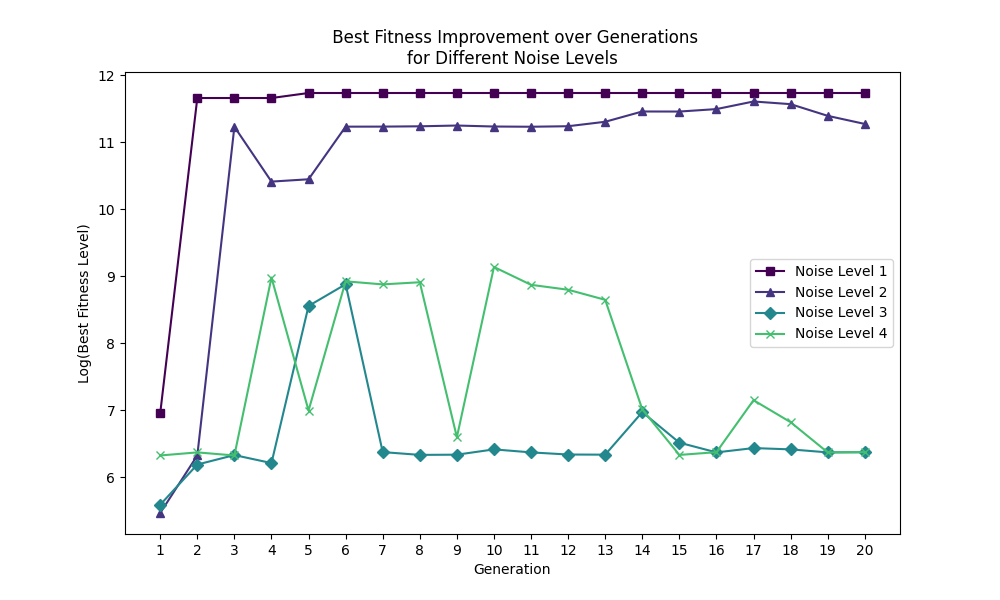}
        \caption{Best fitness (no filtering).}
        \label{fig:best_fitness_no_particle_filter}
    \end{minipage}
\end{figure}

\paragraph{Best Fitness and Robustness.}
Additional plots in Figures~\ref{fig:best_fitness_with_particle_filter} and \ref{fig:best_fitness_no_particle_filter} compare the top-performing controllers each generation. Filtered runs exhibit smoother progression and higher peaks in fitness. Moreover, evolved controllers without filtering tend to fail in higher-noise conditions, highlighting a lack of robustness.

\paragraph{Navigation Paths.}
Qualitative inspection of navigation traces suggests that filtering-based NEAT controllers yield smoother, more direct paths. In contrast, unfiltered controllers exhibit erratic steering when noise intensifies, failing to reach or maintain route consistency.

\subsection{Overall Discussion}

Both experiments confirm that integrating particle filtering into adaptive learning significantly improves outcomes in noisy environments:

\begin{itemize}
    \item \textbf{Q-learning:} Faster convergence, reduced variance, higher final rewards, and increased success rate.
    \item \textbf{NEAT:} Higher fitness scores, smoother evolutionary trends, and more robust controllers under sensor noise.
\end{itemize}

These findings emphasize that accurate state estimation can be a foundational component of RL and evolutionary methods alike, preventing algorithms from “learning the noise” rather than the true dynamics. The next section provides conclusions and explores possible extensions.

\section{Conclusions and Future Directions}
\label{sec:conclusion}

We have presented a framework that integrates particle filtering with two adaptive machine learning methods, Q-learning and NEAT, to address navigation tasks in noisy environments. By refining sensor observations into more accurate state estimates, the particle filter substantially enhances both Q-learning updates and the fitness evaluations of neuroevolved controllers. Experiments on a grid-based domain and a simulated car-navigation problem demonstrated several key findings. First, particle filtering reduces instability and accelerates convergence in Q-learning by providing more robust learning under sensor noise, yielding higher average returns and lower reward variance. Second, accurate state estimates substantially improve evolutionary outcomes in NEAT, enabling higher fitness scores, smoother navigation paths, and greater resilience to elevated noise levels. Third, state estimation is not merely a preprocessing step, but a critical component of adaptive algorithms: the gap between filtered and unfiltered baselines highlights the necessity of reliable state feedback to avoid learning the noise and to achieve significantly more reliable policies or controllers in both reinforcement learning and neuroevolution.

From a broader perspective, these results confirm that advanced filtering techniques can be integral to successful machine learning approaches in real-world applications where sensor data are prone to nontrivial uncertainties. Classical Kalman-based methods may falter when their linear or moderate-noise assumptions fail, whereas particle filtering accommodates nonlinear, non-Gaussian scenarios at the cost of higher computational demands. Recent developments in score-based diffusion models and direct parameter-filtering strategies further suggest that combining generative methods with adaptive learning may yield robust, high-dimensional state estimation at scale.

Looking ahead, several promising directions remain for extending this work. High-dimensional domains and partial observability can challenge naive particle filtering implementations, motivating diffusion-based generative priors or ensemble-based sampling to alleviate sample degeneracy. Real-time adaptation and online parameter estimation can be pursued by unifying direct parameter filtering with Q-learning or NEAT, particularly in environments where system dynamics or noise characteristics shift over time. Additionally, hybrid strategies that integrate particle filtering into model-based RL approaches such as Dreamer or MuZero might merge learned model representations with robust posterior estimation for enhanced planning and control. Overall, our findings reinforce that robust filtering is vital for reinforcement learning and neuroevolution in noisy sensor environments, and emerging generative techniques promise continued improvements in accuracy and scalability for state estimation under uncertainty.

\newpage
\appendix

\section*{Appendix A: Q-Learning Experiment Details}
\label{sec:appendixA}

\subsection*{A.1 Domain Setup and Main Parameters}

\begin{table}[h]
\centering
\caption{Key Domain and Algorithm Parameters for the Q-learning Experiment}
\label{tab:appendixA_tab1}
\begin{tabular}{lp{13cm}}
\hline
\textbf{Aspect} & \textbf{Details} \\
\hline
\textbf{State Domain} & $[0,12]\times[0,12]$ (continuous), discretized into $51\times51$ grid cells\\
\textbf{Action Space} & 8 discrete angles plus speed in $[0.4,1.4]$ \\
\textbf{Initial Position} & $(x_0,y_0)=(2.8,2.8)$ \\
\textbf{Radar Noise} & Gaussian, standard deviation $\sigma = 0.05\,$ (angle domain)\\
\textbf{Particle Filter} & 500 particles; small process noise ($\sigma=0.07$)\\
\textbf{Episodes} & 30,000 total\\
\textbf{Max Steps/Episode} & 100 (terminates if boundary collision or final target reached) \\
\textbf{Q-learning} & $\alpha=0.001$, $\gamma=0.999$, $\epsilon$ decays from $1.0$ to $10^{-5}$ \\
\hline
\end{tabular}
\end{table}

\subsection*{A.2 Wave-Shaped Target Path}

\noindent
\textbf{Path Generation.} 
A sinusoidal curve is constructed from $\,(4,4)\,$ to $\,(11.4,11.4)\,$ with amplitude $2.0$ and frequency $2$, sampled at intervals of approximately $1.1$ in arc length. This results in a sequence of intermediate targets (each of size $0.8\,\times 0.8$) that the agent must sequentially hit. The final target at $(11.4,11.4)$ gives a large terminal reward.

\subsection*{A.3 Reward Structure and Penalties}

\begin{table}[h]
\centering
\caption{Summary of Reward and Penalty Conditions in the Q-learning Experiment}
\label{tab:appendixA_rewards}
\begin{tabular}{lp{13cm}}
\hline
\textbf{Condition} & \textbf{Reward or Penalty} \\
\hline
\textbf{Intermediate Targets} 
& Hitting each wave node yields a guiding reward, \\
& scaled inversely by the distance to the next node. \\[0.3em]
\textbf{Boundary Collisions} 
& Large immediate penalty of $-50{,}000$, ending the episode. \\[0.3em]
\textbf{Idle / Miss Penalty} 
& If no target is hit within 2–3 steps, the reward can reset or \\
& a negative value is imposed proportional to the distance \\
& to the final target. \\[0.3em]
\textbf{Final Target} 
& Large terminal reward (e.g.\ $20^3=8000$) upon successful completion. \\
\hline
\end{tabular}
\end{table}

\clearpage
\section*{Appendix B: NEAT-Based Car Navigation Experiment}
\label{appendix:neat_b}

This appendix details the second experiment, in which a car navigates a track under noisy sensor measurements, aided by a particle filter and trained via NeuroEvolution of Augmenting Topologies (NEAT).

\subsection*{B.1 Environment Setup}

\begin{table}[h]
\centering
\caption{Main Environment Parameters for the NEAT Car Experiment}
\begin{tabular}{lp{13cm}}
\hline
\textbf{Item} & \textbf{Description} \\ \hline
\textbf{Map Dimensions} & $2000 \times 2080$ pixels, with boundary color to detect collisions. \\
\textbf{Car Sprite Size} & $40 \times 40$ pixels. \\
\textbf{Initial Car State} & Position $(x_0,y_0)=(830,920)$; orientation $\theta=0^\circ$; speed $\approx 20$.\\
\textbf{Collision Criterion} & If any corner pixel touches boundary color, car is deemed crashed.\\
\textbf{Time/Step Limit} & Typically 500 steps or until collision.\\
\hline
\end{tabular}
\end{table}

\vspace{-1em}
\subsection*{B.2 Radar Angles and Particle Filtering}

\begin{table}[h]
\centering
\caption{Radar Setup and Noise Modeling}
\begin{tabular}{lp{13cm}}
\hline
\textbf{Aspect} & \textbf{Details} \\ \hline
\textbf{Radar Angles} & Five fixed radars at relative angles $\{-90^\circ,\,-45^\circ,\,0^\circ,\,45^\circ,\,90^\circ\}$ w.r.t.\ car heading.\\
\textbf{Angle Noise} & $\sigma_{\theta}$ added to each radar direction. (Varies in experiments $0\rightarrow 20$.)\\
\textbf{Distance Noise} & $\sigma_{d}$ added to measured distance-to-boundary. (Varies $0\rightarrow 50$.)\\
\textbf{Particle Filter} & One filter per radar, each with $\mathrm{numParticles} \!=\! 30$. 
Updates incorporate \emph{(a)} sensor measurement and \emph{(b)} control-based motion.\\
\hline
\end{tabular}
\end{table}

\subsection*{B.3 NEAT Hyperparameters}

Table~\ref{tab:neat-params} summarizes the key genetic and network parameters used in the NEAT algorithm. 

\begin{table}[h]
\centering
\caption{Key NEAT Parameters for the Car Navigation Task}
\label{tab:neat-params}
\begin{tabular}{lp{13cm}}
\hline
\textbf{Parameter} & \textbf{Value / Notes} \\
\hline
\textbf{Num.~Inputs} & 5 (corresponding to radar-based signals) \\
\textbf{Num.~Outputs} & 4 (\{turn-left, turn-right, slow-down, speed-up\}) \\
\textbf{Population Size} & 30 \\
\textbf{Generations} & 40 (upper bound for training) \\
\textbf{Activation} & \texttt{tanh} (default) \\
\textbf{Elitism} & 3 (best individuals kept) \\
\textbf{Survival Threshold} & 0.2 \\
\textbf{Conn.~Add / Delete Prob} & 0.5 / 0.5 \\
\textbf{Node Add / Delete Prob} & 0.2 / 0.2 \\
\textbf{Compatibility Threshold} & 2.0 \\
\textbf{Weight Mutate Rate} & 0.8 \\
\textbf{Bias Mutate Rate} & 0.7 \\
\hline
\end{tabular}
\end{table}

\end{document}